\documentclass{article}
\usepackage{spconf,amsmath,graphicx}
\usepackage{amssymb}
\usepackage{makecell}
\usepackage{color}
\usepackage{multirow}
\usepackage[ruled]{algorithm2e}

\title{DMCNN: Dual-Domain Multi-scale Convolutional neural Network 
for Compression Artifacts Removal\thanks{$\copyright$ 2018 IEEE}}
\name{Xiaoshuai Zhang, Wenhan Yang, Yueyu Hu, Jiaying Liu
\sthanks{Corresponding author. This work was supported by
National Natural Science Foundation of China under contract
No. 61772043, CCF-Tencent Open Research Fund and in part
by the PKU-NTU Joint Research Institute (JRI) sponsored
by a donation from the Ng Teng Fong Charitable Foundation.}}
\address{Institute of Computer Science and Technology, Peking University, Beijing, China}
\begin{document}
%\ninept
%
\maketitle
\vspace{-0.2cm}
\begin{abstract}
JPEG is one of the most commonly used standards among lossy image compression
methods. However, JPEG compression inevitably introduces various kinds of
artifacts, especially at high compression rates, which could greatly affect
the Quality of Experience (QoE). Recently, convolutional neural network (CNN)
based methods have shown excellent performance for removing the JPEG artifacts.
Lots of efforts have been made to deepen the CNNs and extract deeper features,
while relatively few works pay attention to the receptive field of the network.
In this paper, we illustrate that the quality of output images can be
significantly improved by enlarging the receptive fields in many cases.
One step further, we propose a Dual-domain Multi-scale CNN (DMCNN) to
take full advantage of redundancies on both the pixel and DCT domains.
Experiments show that DMCNN sets a new state-of-the-art for the task of
JPEG artifact removal.
\end{abstract}
\begin{keywords}
Compression Artifacts Removal, Image Restoration, JPEG, Convolutional Neural
Network
\end{keywords}
\vspace{-0.2cm}
\section{Introduction}
\label{sec:intro}
Nowadays, lossy image compression methods (\textit{e.g.} JPEG, HEVC-MSP and WebP)
have been used extensively for image storage and transmission.
These methods typically shrink parts of image information by quantization
and approximation, so that higher compression rates can be reached. These
methods can usually reduce the bit-rate greatly but still maintain satisfactory
visual quality by taking advantage of the limitation of the
human visual system. But as the compression rate increases, these methods
tend to introduce undesirable artifacts such as
blocking, ringing, and banding. These artifacts severely degrade
the user experience.

In this paper, we examine the degradation of JPEG-compressed images.
Typically, a JPEG compressor converts an image of RGB color space into
the YCbCr color space. The chroma channels (namely Cb and Cr) are downsampled
by the factor of 2. Then, the image is partitioned into $8 \times 8$ blocks and
the block-wise 2D Discrete Cosine Transform (DCT) is performed. After DCT,
the top left items in each $8 \times 8$ block are low-frequency components,
representing the overall features such as the average luminance. The bottom right
items are high-frequency components, representing local features such as
textures and details. Next, quantization is applied on each of
64 DCT coefficients. As human eyes are not so good at distinguishing
high frequency brightness variation, quantization intervals are typically
much larger on high-frequency components than low-frequency ones.
Noting that the quantization step is the culprit for various kinds of artifacts
such as the blocking artifacts within the boundaries of each $8 \times 8$
DCT block, the ringing artifacts around sharp edges, and the noticable
banding effects over the image. As a matter of fact, these kinds of artifacts
can be commonly seen on other transform-based methods.

Many methods have been proposed to improve the quality of JPEG-compressed
images. Traditional filter-based methods~\cite{foi2007pointwise,dabov2007image}
pay attention to general images denoising. Others apply the sparse coding (SC)
to restore the compressed images \cite{chang2014reducing,liu2015data}.
These methods generally produce sharper images given a compressed input,
but they are usually too slow and their results are often accompanied
with additional artifacts.

With the rapid development of deep neural networks, multiple CNN-based methods
have been used in low-level image processing,
including denoising~\cite{zhang2017beyond,zhang2018dynamically},
super-resolution~\cite{Dong2016SRCNN,Yang2018MALDEC,YANG201879},
video compression~\cite{yueyu_2018_DCC,Sifeng_2018_DCC},
rain removal~\cite{Fu_2017_CVPR,Qian_2018_CVPR,Jiaying_2018_CVPR}.
Specifically for compression artifacts removal, Dong \textit{et al.}
\cite{dong2015compression} first introduce a CNN-based method and
the proposed ARCNN set a good practice for the following low-level CNN-based
methods including DnCNN \cite{zhang2017beyond}, CAS-CNN \cite{cavigelli2017cas},
and MemNet \cite{tai2017memnet}.
However, these methods usually work on pixel domain only and do not
incorporate much JPEG prior knowledge. More recently, a dual-domain
CNN-based model DDCN is proposed in  \cite{guo2016building}.
The model sucessfully combines DCT-domain prior and the power of
the CNN, thus achieves impressive performance.

\begin{figure}[htb]
  \begin{minipage}[b]{1.0\linewidth}
    \centering
    \centerline{\includegraphics[width=8.2cm]{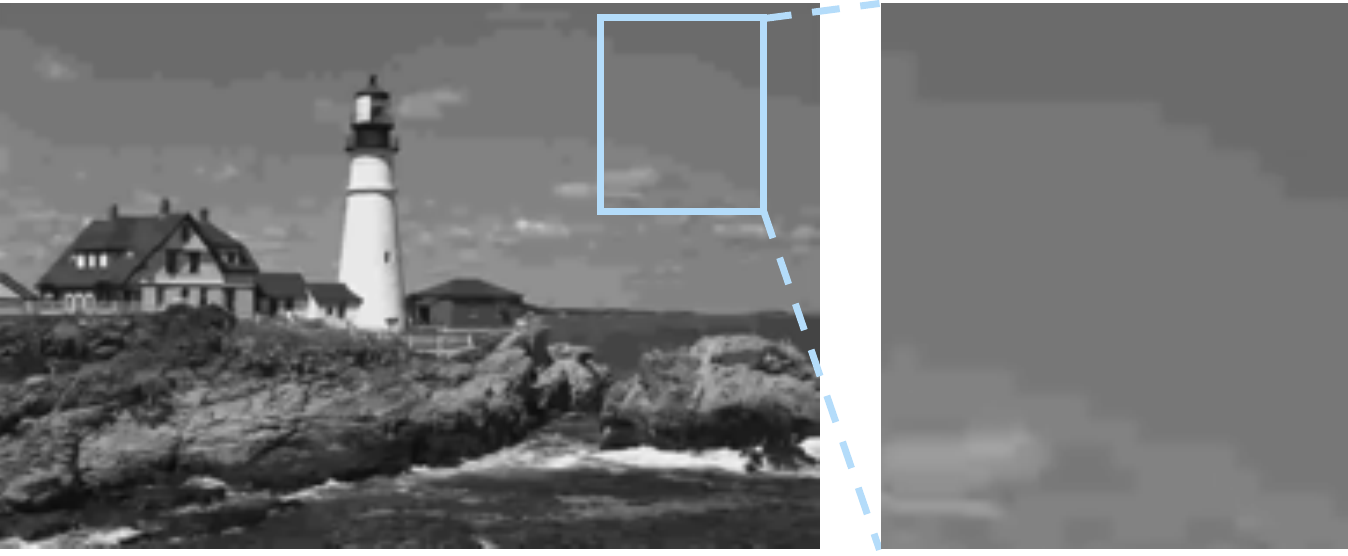}}
    % \vspace{2.0cm}
    \caption{The banding effects can still be clearly seen
    after the process of ARCNN (QF=10).}\medskip
    \label{fig:f1}
    \vspace{-0.5cm}
  \end{minipage}
\end{figure}

However, a common weakness of all these CNN-based methods is that
the receptive fields of their models are too small, and these models are
usually trained on mini image patches (\textit{e.g.} $35 \times 35$ for ARCNN,
$49 \times 49$ for DnCNN, $55 \times 55$ for DDCN), so that only a
small range of information (\textit{i.e.} neighbor pixels) are
taken into consideration when performing the restoration. Noting that
local information is sometimes not enough to remove all the artifacts,
especially the banding effects, which can appear at a large scale on the image.
As shown in Fig. \ref{fig:f1}, due to the quantization step of JPEG,
the sky in the image is split into multiple bandings, and ARCNN fails to
recover the bandings because of small receptive fields.

To eliminate the banding effects, three efforts have been made to
enlarge the receptive fields and enable our model with
the ability of extracting global features:
(1) Auto-encoder style architecture; (2) Dilated convolutions;
(3) Multi-scale loss. Moreover, as validated by DDCN, redundancies in
the DCT domain can be effectively utilized. We also adopt a DCT domain branch
to enhance the performance of the proposed model.

Our major contribution is to propose an end-to-end CNN with
large receptive fields to exploit dual-domain multi-scale features. In order to
train the proposed DMCNN more effectively, a modified version of
residual learning as well as other training techniques have been utilized.
The evaluations on the BSDS500 and the LIVE1 dataset have shown that our work
is the current state-of-the-art among all CNN-based JPEG artifact removal networks.
\vspace{-0.2cm}

\section{Dual-Domain Multi-scale
Convolutional Neural Network (DMCNN)}
\label{sec:model}
\vspace{-0.2cm}
   
\begin{figure*}[htb]
  \vspace{-0.6cm}
  \begin{minipage}[b]{1.0\linewidth}
    \centering
    \centerline{\includegraphics[width=16cm]{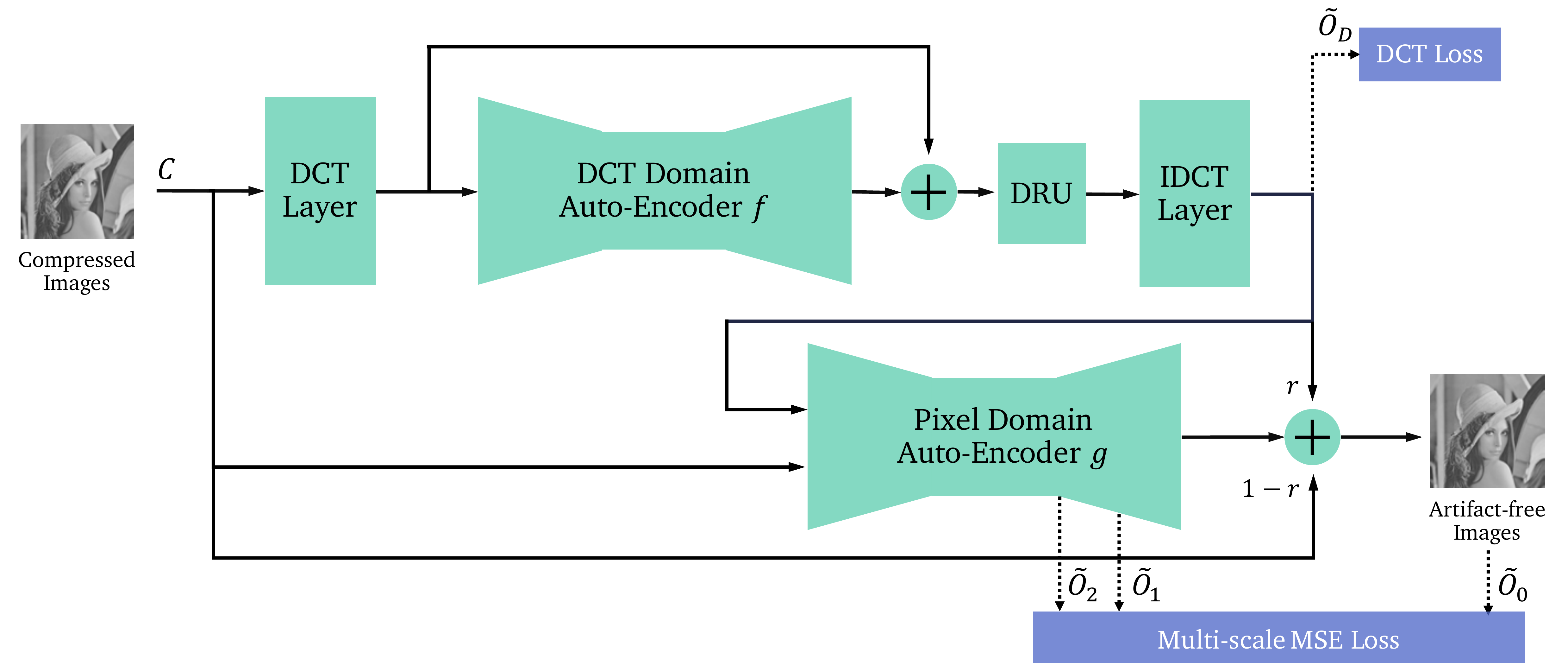}}
    \vspace{-0.2cm}
    \caption{The architecture of Dual-domain Multi-scale
    Convolutional Network (DMCNN).}\medskip
    \label{fig:f2}
  \end{minipage}
  \vspace{-1.0cm}
\end{figure*}

The architecture of our proposed DMCNN is in Fig. \ref{fig:f2}.
The model is mainly composed of two similar auto-encoder style networks
working on pixel and DCT domains, respectively. The input image is processed
by DCT branch first, then passed into the pixel domain branch. The final
restoration result is the weighted sum of the input, the DCT branch estimation
and the pixel branch estimation.
\vspace{-0.2cm}

\subsection{Auto-Encoder}
\label{ssec:ae}

The auto-encoder is an efficient way to learn a representation for given data,
and typically used for the purpose of dimensionality reduction.
An auto-encoder consists of two parts, the encoder $E$ and
the decoder $D$, so that:

\begin{equation}
  \begin{aligned}
    E&:\mathcal{X} \rightarrow \mathcal{F}, \\
    D&:\mathcal{F} \rightarrow \mathcal{X}, \\
    \displaystyle E,D =\ &
    {\underset{E,D}
    {\operatorname {arg\,min} }}\,\|x-(D \circ E )x\|^{2},
  \end{aligned}
\end{equation}

where $\mathcal{X}$ denotes the set of the to-be-compressed data, $\mathcal{F}$
denotes the feature space and $\circ$ is the composition operator.
Typically, an auto-encoder is trained with identical
image pairs $(x, x)\in \mathcal{X}\times\mathcal{X}$, so that $E$ could learn to extract the best
representation of the data in $\mathcal{F}$.

We adopt an auto-encoder style network here as a generative model. Given
a compressed image, the encoder $E$ is expected to extract
artifact-free features robustly, and then the image is restored from these
clean features by decoder $D$.

In order to learn both local and global features, shortcuts are linked between
symmetric layers. Also, residual learning strategy is employed.
To stress the information learnt form the DCT domain,
we add a shortcut with parameter $r$ from the DCT-branch into the final result.
Given an input image $C$, the intermediate estimation of the DCT branch
$\tilde O_D$ and the final output $\tilde O_0$ can be fomulated as:
\begin{equation}
  \begin{aligned}
    \tilde O_D &= \mathcal{D}^{-1}([f(\mathcal{D}(C))]_\text{DRU}),\\
    \tilde O_0 &= g(C, \tilde O_D) + r \tilde O_D + (1-r) C,
  \end{aligned}
\end{equation}
where $\mathcal{D}$ and $\mathcal{D}^{-1}$ stand for the process of
$8 \times 8$ block-wise DCT and inverse DCT (IDCT) respectively. $f(\cdot)$
and $g(\cdot)$ denote the processes of the DCT domain auto-encoder and
the pixel domain auto-encoder, respectively. $[\cdot]_\text{DRU}$ denotes
the DCT Rectify Unit stated later, and $r$ is a learnable parameter of
the residual addition module.
\vspace{-0.2cm}

\subsection{Dilated Convolution}
\label{ssec:diconv}

The dilated convolution is a kind of convolution with pre-defined gaps,
it is first named in \cite{YuKoltun2016}. Consider
an input image $I$ as a discrete function $I:\mathbb{Z}^2 \rightarrow\mathbb{R}$
and a convolution kernel $k$ shaped $(2r+1)\times(2r+1)$ as a discrete function
$k:\Omega_r\rightarrow\mathbb{R}$, where $\Omega_r=[-r,r]^2\cap\mathbb{Z}^2$.
The discrete convolution operator $*_d$ with dilation factor $d$ can
be defined as:

\begin{equation}
  (I*_dk)(\mathbf{p}) =
  \sum_{\mathbf{s}+d\mathbf{t}=\mathbf{p}}{I(\mathbf{s})k(\mathbf{t})},
\end{equation}
where $\mathbf{p}, \mathbf{s}, \mathbf{t} \in \mathbb{Z}^2$ are 2D indices.

Unlike normal convolutions, the receptive field of $n$ combined
dilated convolutions can reach $(2^{n-1}-1) \times (2^{n-1}-1)$ when
dilation factors are set to $1, 2, 4, ..., 2^{n-1}$, respectively.
The dilated convolution has been widely used in other vision tasks
\cite{chen2016deeplab,wang2017large} and shown considerable gain,
but to the best of our knowledge, has not been used in the
task of artifacts removal. In our model, the dilated convolutional layers
with dilation factors 2, 4, 8 are used in the middle of the auto-encoder,
which aim to enlarge the receptive field further.

With the combination of auto-encoder style architecture and
dilated convolutions, the receptive field of our proposed DMCNN reaches
$145 \times 145$, which is about 58 times larger than the ARCNN
($19 \times 19$), and 13 times larger than the DnCNN and
the DDCN ($41 \times 41$).
\vspace{-0.2cm}

\subsection{DCT Rectify Unit (DRU)}
\label{ssec:dru}

As stated earlier in the introduction, the main cause of the
JPEG compression artifacts is the step of quantization. The quantization step
in each of the $8 \times 8$ compression blocks can be formulated as follows:
\begin{equation}
  \begin{aligned}
    C^\text{DCT}(\mathbf{p}) = \text{round}(O^\text{DCT}(\mathbf{p})
    / Q(\mathbf{p})) * Q(\mathbf{p}),
  \end{aligned}
\end{equation}
where $C^\text{DCT}$ is the quantized DCT block, $O^\text{DCT}$ is
the original DCT block, $Q$ is the quantization table, and
$\mathbf{p} \in \mathbb{Z}^2$ is a 2D index. $/$ here denotes
the element-wise division.

From $(4)$, it can be easily seen that the estimated $\tilde O^\text{DCT}$
should meet the requirement:
\begin{equation}
  C^\text{DCT} - Q/2 \le \tilde O^\text{DCT} \le C^\text{DCT} + Q/2.
\end{equation}
So like \cite{guo2016building} we employ a DCT Rectify Unit (DRU) to constraint
the value of DCT block elements, where values out of the range will be cropped.
A slight difference is that we drop the leaky slope $\alpha$ in their proposed
unit, as no gain can be observed with it. Our DRU can be formulated as:

\footnotesize
\begin{equation}
  [X]_\text{DRU}(\mathbf{p})=
    \begin{cases}
      C^\text{DCT}(\mathbf{p}) - Q(\mathbf{p}) / 2,
      & X(\mathbf{p})<C^\text{DCT}(\mathbf{p}) - Q(\mathbf{p}) / 2,  \\
      C^\text{DCT}(\mathbf{p}) + Q(\mathbf{p}) / 2, 
      & X(\mathbf{p})>C^\text{DCT}(\mathbf{p}) + Q(\mathbf{p}) / 2,  \\
      X(\mathbf{p}),
      & \text{otherwise}.
    \end{cases}
\end{equation}
\normalsize

\vspace{-0.3cm}
\subsection{Multi-scale DCT-Embedded Loss}
\label{ssec:mmse}

As is pointed out by previous works \cite{dong2014learning}
 \cite{dong2015compression}, ``deeper is not better'' in certain
low-level tasks. The reason is that deeper neural networks are usually harder
to train due to the gradient vanishing. We try to address this problem by
redesigning the loss function of the model.

A multi-scale loss is adopted to extract features at different scales.
More specifically, features are extracted from different
deconvolutional layers of the pixel domain decoder, and
scaled images are expected to be reconstructed from these features.
By adopting the multi-scale loss, we explicitly guide our network to learn
features at different scales. 

We also add a DCT loss to train the DCT branch more effectively. Finally,
our loss function can be stated as:
\begin{equation}
  \begin{aligned}
    &\mathcal{L}({\{\tilde O\}}_{i=0}^2,\tilde O_D , \{O\}_{i=0}^2) \\
    &= \mathcal{L}_{\text{MMSE}}(\{\tilde O\}_{i=0}^2, \{O\}_{i=0}^2)
    + \lambda \mathcal{L}_{\text{DCT}}(\tilde O_D , O_0) \\
    &= \sum_{i=0}^{2}{\theta^i\text{MSE}(\tilde O_i, O_i)} 
    + \lambda \text{MSE}(\tilde O_D, O_0),
  \end{aligned}
\end{equation}
where ${\{\tilde O\}}_{i=0}^2$ are estimations from pixel domain auto-encoder
at different scales, $\{O\}_{i=0}^2$ are original images at different scales,
and $\tilde O_D$ is the intermeidate estimation of the DCT branch.
Hyper-parameters $\lambda$ and $\theta$ are used to adjust the weights of
each loss, they should typically be in the range of $[0, 1]$.
MSE($\cdot$,$\cdot$) denotes the mean squared error loss.
\vspace{-0.2cm}

\section{Experiments}
\vspace{-0.1cm}
\label{sec:exp}
\subsection{Implement Details}
\label{ssec:detail}

\begin{figure*}[htb]
  \vspace{-0.5cm}
  \begin{minipage}[b]{1.0\linewidth}
    \centering
    \centerline{\includegraphics[width=18cm]{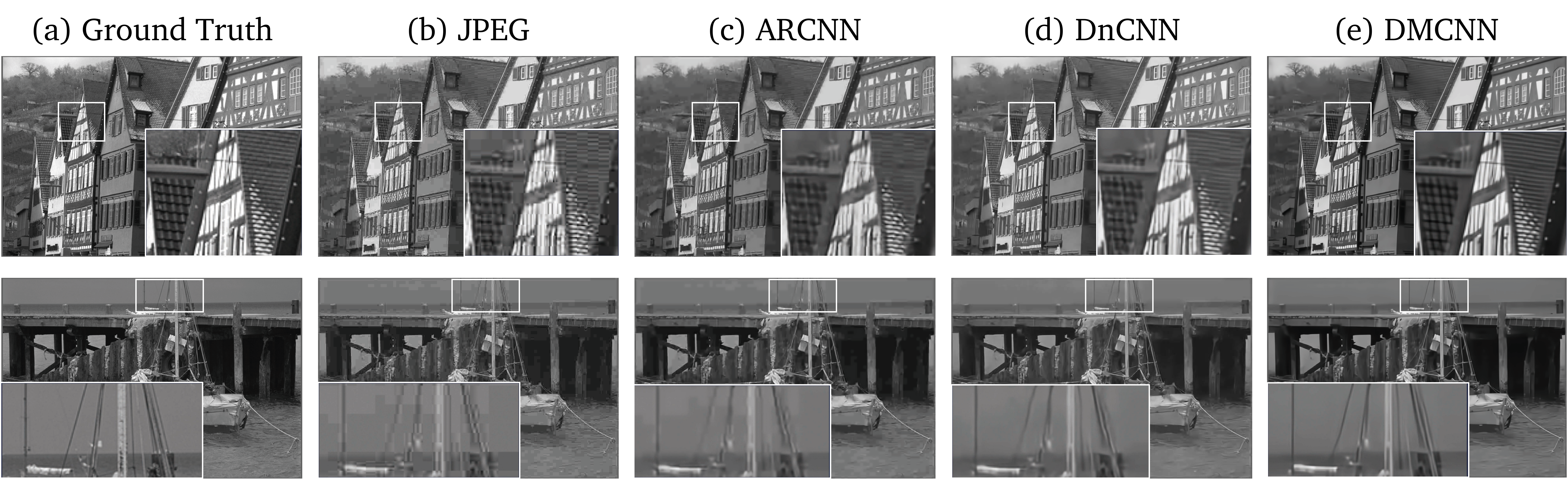}}
    % \vspace{2.0cm}
    \caption{Visual comparisons between different algorithms with QF=10.
    Zooming-in the figure will provide a better look at the restoration quality.}\medskip
    \label{fig:f3}
    \vspace{-0.7cm}
  \end{minipage}
\end{figure*}

\textbf{Datasets.\ }
In all experiments, we employ the ImageNet \cite{ILSVRC15} for training.
The LIVE1 dataset and the testing set of BSDS500 are used for evaluation.
All training and evaluation processes are done on gray-scale images
(the Y channel of YCbCr space). The PIL module of python is applied to
generate JPEG-compressed images. The module produces numerically identical images
as the commonly used MATLAB JPEG encoder after setting the quantization tables manually.
\vspace{-0.1cm}
\\\\
\textbf{Parameter Settings.\ }
The hyper-parameters $\lambda$ and $\theta$ are set to $0.9$ and $0.618$,
respectively. The parameter $r$ of final residual links is initialized
to $0.5$. The DCT and IDCT layers are fixed and initialized with the
corresponding DCT matrix coefficients. Leaky slopes are initialized
to $0.1$ for PReLUs. The depth of the pixel and DCT domain auto-encoders
are 15 and 9, respectively.
\vspace{-0.1cm}
\\\\
\textbf{Training Details.\ }
Adam optimizer with initial learning rate 0.001 is used for training.
The learning rate is scaled down by the factor of 3 when
the validation loss stops decreasing. The batch size is set to 80.
Training pairs are dynamically generated from the training set. The sizes of
training patches are not fixed. As an easy-to-hard transfer, we first train
our model on pairs generated with quality factor (QF) of 20 and patch size
of $56 \times 56$. Then we gradually increase the patch size
till $224 \times 224$. After full convergence, the model
dedicated to QF10 is trained based on the previous QF20 model.  \\\\
\vspace{-1.1cm}
\subsection{Objective Comparisons}
\vspace{-0.2cm}
\label{ssec:bsds500}
To have overall comparisons, we calculate the mean PSNR,
SSIM \cite{wang2004image}, and PSNR-B \cite{yim2011quality} on the
two datasets. We compare to recent state-of-the-arts including
pixel domain methods -- ARCNN and CAS-CNN, dual domain method
-- DDCN and general frameworks -- TNRD \cite{chen2017trainable}
and DnCNN\cite{zhang2017beyond}.

The quantitative results are shown in Table.\ref{tab:t1} and
Table.\ref{tab:t2}. Generally, our
proposed model DMCNN outperforms all the other methods on all evaluated
datasets, QFs and metrics. Specifically, our model far surpasses all
pixel domain methods and general frameworks. Also, considerable gains
can be observed compared to the dual domain method DDCN.

\begin{table}[!htb]
  \vspace{-0.5cm}
  \caption{The quantitative results on LIVE1.}\medskip
  \vspace{-0.1cm}
  \centering
  \begin{tabular}{c|c|c|c|c}
  \Xhline{2\arrayrulewidth}{QF}
  &Method                            &PSNR(dB)       &SSIM           &PSNR-B(dB)    \\ 
  \hline
  \multirow{6}{*}{10}
  &JPEG                              &27.77          &0.791          &25.33         \\
  &ARCNN \cite{dong2015compression}  &29.13          &0.823          &28.74         \\
  &TNRD \cite{chen2017trainable}     &29.24          &0.825          &28.90         \\
  &DnCNN-3 \cite{zhang2017beyond}    &29.27          &0.825          &28.98         \\
  &CAS-CNN \cite{cavigelli2017cas}   &29.44          &0.833          &29.19         \\
  &DMCNN                             &\textbf{29.73} &\textbf{0.842} &\textbf{29.55}\\
  \hline
  \multirow{6}{*}{20}
  &JPEG                              &30.07          &0.868          &27.57         \\
  &ARCNN \cite{dong2015compression}  &31.40          &0.890          &30.69         \\
  &TNRD \cite{chen2017trainable}     &31.52          &0.892          &30.88         \\
  &DnCNN-3 \cite{zhang2017beyond}    &31.62          &0.894          &30.89         \\
  &CAS-CNN \cite{cavigelli2017cas}   &31.70          &0.895          &30.88         \\
  &DMCNN                             &\textbf{32.09} &\textbf{0.905} &\textbf{31.32}\\
  \Xhline{2\arrayrulewidth}
  \multicolumn{5}{c}{}\\
\end{tabular}
\label{tab:t1}
\vspace{-0.8cm}
\end{table}
\begin{table}[!htb]
  \vspace{-0.2cm}
  \caption{The quantitative results on BSDS500 testing set.}\medskip
  \vspace{-0.1cm}
  \centering
  \begin{tabular}{c|c|c|c|c}
  \Xhline{2\arrayrulewidth}{QF}
  &Method                            &PSNR(dB)       &SSIM           &PSNR-B(dB)    \\ 
  \hline
  \multirow{6}{*}{10}
  &JPEG                              &27.80          &0.788          &25.10         \\ 
  &ARCNN \cite{dong2015compression}  &29.10          &0.820          &28.73         \\
  &TNRD \cite{chen2017trainable}     &29.16          &0.823          &28.81         \\
  &DnCNN-3 \cite{zhang2017beyond}    &29.17          &0.823          &28.91         \\
  &DDCN \cite{guo2016building}       &29.59          &0.838          &29.18         \\
  &DMCNN                             &\textbf{29.67} &\textbf{0.840} &\textbf{29.33}\\
  \hline
  \multirow{6}{*}{20}
  &JPEG                              &30.05          &0.867          &27.22         \\ 
  &ARCNN \cite{dong2015compression}  &31.28          &0.885          &30.55         \\
  &TNRD \cite{chen2017trainable}     &31.41          &0.889          &30.83         \\
  &DnCNN-3 \cite{zhang2017beyond}    &31.50          &0.891          &30.85         \\
  &DDCN \cite{guo2016building}       &31.88          &0.900          &31.10         \\
  &DMCNN                             &\textbf{31.98} &\textbf{0.904} &\textbf{31.29}\\
  \Xhline{2\arrayrulewidth}
  \end{tabular}
  \label{tab:t2}
  \vspace{-0.6cm}
\end{table}
\vspace{-0.2cm}
\subsection{Subjective Comparisons}
\vspace{-0.15cm}
\label{ssec:qual}
For subjective comparisons, some restored images of different approches on
the LIVE1 dataset have been presented. As can be seen in Fig. \ref{fig:f3},
the results of DMCNN are more visually pleasing. Due to the large
receptive fields, our model is able to handle quantization banding effects
as well as recover lost details using regional patterns.
More experiment results are available on our project page
\footnote{http://i.buriedjet.com/projects/DMCNN/}.

\vspace{-0.3cm}
\section{Conclusions}
\vspace{-0.3cm}
\label{sec:conc}
In this paper, we introduce a novel network based on dual-domain auto-encoders,
named DMCNN. By applying dilated convolutional layers and multi-scale loss,
our model is able to extract global information and eliminate
JPEG compression artifacts effectively.

% Below is an example of how to insert images. Delete the ``\vspace'' line,
% uncomment the preceding line ``\centerline...'' and replace ``imageX.ps''
% with a suitable PostScript file name.
% -------------------------------------------------------------------------

% To start a new column (but not a new page) and help balance the last-page
% column length use \vfill\pagebreak.
% -------------------------------------------------------------------------
%\vfill
%\pagebreak

% List and number all bibliographical references at the end of the
% paper. The references can be numbered in alphabetic order or in
% order of appearance in the document. When referring to them in
% the text, type the corresponding reference number in square
% brackets as shown at the end of this sentence  \cite{C2}. An
% additional final page (the fifth page, in most cases) is
% allowed, but must contain only references to the prior
% literature.

% References should be produced using the bibtex program from suitable
% BiBTeX files (here: strings, refs, manuals). The IEEEbib.bst bibliography
% style file from IEEE produces unsorted bibliography list.
% -------------------------------------------------------------------------
\bibliographystyle{IEEEbib}
\small
\bibliography{refs}

\begin{thebibliography}{10}

\bibitem{foi2007pointwise}
Alessandro Foi, Vladimir Katkovnik, and Karen Egiazarian,
\newblock ``Pointwise shape-adaptive dct for high-quality denoising and
  deblocking of grayscale and color images,''
\newblock {\em IEEE Transactions on Image Processing}, vol. 16, no. 5, pp.
  1395--1411, 2007.

\bibitem{dabov2007image}
Kostadin Dabov, Alessandro Foi, Vladimir Katkovnik, and Karen Egiazarian,
\newblock ``Image denoising by sparse 3-d transform-domain collaborative
  filtering,''
\newblock {\em IEEE Transactions on Image Processing}, vol. 16, no. 8, pp.
  2080--2095, 2007.

\bibitem{chang2014reducing}
Huibin Chang, Michael~K Ng, and Tieyong Zeng,
\newblock ``Reducing artifacts in jpeg decompression via a learned
  dictionary,''
\newblock {\em IEEE Transactions on Signal Processing}, vol. 62, no. 3, pp.
  718--728, 2014.

\bibitem{liu2015data}
Xianming Liu, Xiaolin Wu, Jiantao Zhou, and Debin Zhao,
\newblock ``Data-driven sparsity-based restoration of jpeg-compressed images in
  dual transform-pixel domain.,''
\newblock in {\em Proceedings of the IEEE International Conference on Computer
  Vision and Pattern Recognition}, 2015, vol.~1, p.~5.

\bibitem{zhang2017beyond}
Kai Zhang, Wangmeng Zuo, Yunjin Chen, Deyu Meng, and Lei Zhang,
\newblock ``Beyond a {Gaussian} denoiser: Residual learning of deep {CNN} for
  image denoising,''
\newblock {\em IEEE Transactions on Image Processing}, vol. 26, no. 7, pp.
  3142--3155, 2017.

\bibitem{zhang2018dynamically}
Xiaoshuai Zhang, Yiping Lu, Jiaying Liu, and Bin Dong,
\newblock ``Dynamically unfolding recurrent restorer: A moving endpoint control
  method for image restoration,''
\newblock {\em arXiv preprint arXiv:1805.07709}, 2018.

\bibitem{Dong2016SRCNN}
Chao Dong, Chen~Change Loy, Kaiming He, and Xiaoou Tang,
\newblock ``Image super-resolution using deep convolutional networks,''
\newblock {\em {IEEE} Transactions on Pattern Analysis and Machine
  Intelligence}, vol. 38, no. 2, pp. 295--307, Feb 2016.

\bibitem{Yang2018MALDEC}
Wenhan Yang, Sifeng Xia, Jiaying Liu, and Zongming Guo,
\newblock ``Reference guided deep super-resolution via manifold localized
  external compensation,''
\newblock {\em {IEEE} Transactions on Circuits and Systems for Video
  Technology}, pp. 1--1, 2018.

\bibitem{YANG201879}
Wenhan Yang, Jiashi Feng, Guosen Xie, Jiaying Liu, Zongming Guo, and Shuicheng
  Yan,
\newblock ``Video super-resolution based on spatial-temporal recurrent residual
  networks,''
\newblock {\em Computer Vision and Image Understanding}, vol. 168, pp. 79 --
  92, 2018.

\bibitem{yueyu_2018_DCC}
Yueyu Hu, Wenhan Yang, Sifeng Xia, Wen-Huang Cheng, and Jiaying Liu,
\newblock ``Enhanced intra prediction with recurrent neural network in video
  coding,''
\newblock in {\em Proceedings of Data Compression Conference}, March 2018.

\bibitem{Sifeng_2018_DCC}
Sifeng Xia, Wenhan Yang, Yueyu Hu, Siwei Ma, and Jiaying Liu,
\newblock ``A group variational transformation neural network for fractional
  interpolation of video coding,''
\newblock in {\em Proceedings of Data Compression Conference}, March 2018.

\bibitem{Fu_2017_CVPR}
Xueyang Fu, Jiabin Huang, Delu Zeng, Yue Huang, Xinghao Ding, and John Paisley,
\newblock ``Removing rain from single images via a deep detail network,''
\newblock in {\em Proceedings of the IEEE International Conference on Computer
  Vision and Pattern Recognition}, July 2017.

\bibitem{Qian_2018_CVPR}
Rui Qian, Robby~T. Tan, Wenhan Yang, Jiajun Su, and Jiaying Liu,
\newblock ``Attentive generative adversarial network for raindrop removal from
  a single image,''
\newblock in {\em Proceedings of the IEEE International Conference on Computer
  Vision and Pattern Recognition}, June 2018.

\bibitem{Jiaying_2018_CVPR}
Jiaying Liu, Wenhan Yang, Shuai Yang, and Zongming Guo,
\newblock ``Erase or fill? deep joint recurrent rain removal and reconstruction
  in videos,''
\newblock in {\em Proceedings of the IEEE International Conference on Computer
  Vision and Pattern Recognition}, June 2018.

\bibitem{dong2015compression}
Chao Dong, Yubin Deng, Chen Change~Loy, and Xiaoou Tang,
\newblock ``Compression artifacts reduction by a deep convolutional network,''
\newblock in {\em Proceedings of the IEEE International Conference on Computer
  Vision}, 2015, pp. 576--584.

\bibitem{cavigelli2017cas}
Lukas Cavigelli, Pascal Hager, and Luca Benini,
\newblock ``Cas-cnn: A deep convolutional neural network for image compression
  artifact suppression,''
\newblock in {\em Proceedings of the International Joint Conference on Neural
  Networks (IJCNN)}. IEEE, 2017, pp. 752--759.

\bibitem{tai2017memnet}
Ying Tai, Jian Yang, Xiaoming Liu, and Chunyan Xu,
\newblock ``Memnet: A persistent memory network for image restoration,''
\newblock in {\em Proceedings of the IEEE Conference on Computer Vision and
  Pattern Recognition}, 2017, pp. 4539--4547.

\bibitem{guo2016building}
Jun Guo and Hongyang Chao,
\newblock ``Building dual-domain representations for compression artifacts
  reduction,''
\newblock in {\em Proceedings of the European Conference on Computer Vision}.
  Springer, 2016, pp. 628--644.

\bibitem{YuKoltun2016}
Fisher Yu and Vladlen Koltun,
\newblock ``Multi-scale context aggregation by dilated convolutions,''
\newblock in {\em Proceedings of the International Conference on Learning
  Representations}, 2016.

\bibitem{chen2016deeplab}
Liang-Chieh Chen, George Papandreou, Iasonas Kokkinos, Kevin Murphy, and Alan~L
  Yuille,
\newblock ``Deeplab: Semantic image segmentation with deep convolutional nets,
  atrous convolution, and fully connected crfs,''
\newblock {\em arXiv preprint arXiv:1606.00915}, 2016.

\bibitem{wang2017large}
Qiang Wang, Huijie Fan, Yang Cong, and Yandong Tang,
\newblock ``Large receptive field convolutional neural network for image
  super-resolution,''
\newblock in {\em Proceedings of 2017 IEEE International Conference on Image
  Processing}. IEEE, 2017, pp. 958--962.

\bibitem{dong2014learning}
Chao Dong, Chen~Change Loy, Kaiming He, and Xiaoou Tang,
\newblock ``Learning a deep convolutional network for image super-resolution,''
\newblock in {\em Proceedings of the European Conference on Computer Vision}.
  Springer, 2014, pp. 184--199.

\bibitem{ILSVRC15}
Olga Russakovsky, Jia Deng, Hao Su, Jonathan Krause, Sanjeev Satheesh, Sean Ma,
  Zhiheng Huang, Andrej Karpathy, Aditya Khosla, Michael Bernstein,
  Alexander~C. Berg, and Li~Fei-Fei,
\newblock ``{ImageNet Large Scale Visual Recognition Challenge},''
\newblock {\em International Journal of Computer Vision (IJCV)}, vol. 115, no.
  3, pp. 211--252, 2015.

\bibitem{wang2004image}
Zhou Wang, Alan~C Bovik, Hamid~R Sheikh, and Eero~P Simoncelli,
\newblock ``Image quality assessment: from error visibility to structural
  similarity,''
\newblock {\em IEEE Transactions on Image Processing}, vol. 13, no. 4, pp.
  600--612, 2004.

\bibitem{yim2011quality}
Changhoon Yim and Alan~Conrad Bovik,
\newblock ``Quality assessment of deblocked images,''
\newblock {\em IEEE Transactions on Image Processing}, vol. 20, no. 1, pp.
  88--98, 2011.

\bibitem{chen2017trainable}
Yunjin Chen and Thomas Pock,
\newblock ``Trainable nonlinear reaction diffusion: A flexible framework for
  fast and effective image restoration,''
\newblock {\em IEEE Transactions on Pattern Analysis and Machine Intelligence},
  vol. 39, no. 6, pp. 1256--1272, 2017.

\end{thebibliography}

\end{document}